\documentclass[conference]{IEEEtran}

\IEEEoverridecommandlockouts                              

\usepackage{cite}
\usepackage{balance}
\usepackage{amsmath,amssymb,amsfonts}
\usepackage{float}
\usepackage{booktabs} 
\usepackage{placeins}
\makeatletter
\g@addto@macro\normalsize{%
  \setlength\abovedisplayskip{10pt}
  \setlength\belowdisplayskip{10pt}
  \setlength\abovedisplayshortskip{10pt}
  \setlength\belowdisplayshortskip{10pt}
}
\usepackage{algorithmic}
\usepackage{graphicx}
\usepackage{subcaption}
\usepackage{breqn}
\usepackage{textcomp}
\usepackage{url}
\usepackage{comment}
\usepackage{xcolor}

\title{\LARGE \bf
Motion Segmentation from a Moving Monocular Camera
}

\author{Yuxiang Huang$^{1}$ and John Zelek$^{1}$
\thanks{$^{1}$Yuxiang Huang and $^{1}$John Zelek are with Systems Design Engineering Department, University of Waterloo, 
        Waterloo, N2L 3G1, Canada.
        {\tt\small \{yuxiang.huang, jzelek\}@uwaterloo.ca}}%
}

\begin{document}

\maketitle
\thispagestyle{empty}
\pagestyle{empty}

\begin{abstract}

Identifying and segmenting moving objects from a moving monocular camera is difficult when there is unknown camera motion, different types of object motions and complex scene structures. To tackle these challenges, we take advantage of two popular branches of monocular motion segmentation approaches: point trajectory based and optical flow based methods, by synergistically fusing these two highly complementary motion cues at object level. By doing this, we are able to model various complex object motions in different scene structures at once, which has not been achieved by existing methods. We first obtain object-specific point trajectories and optical flow mask for each common object in the video, by leveraging the recent foundational models in object recognition, segmentation and tracking. We then construct two robust affinity matrices representing the pairwise object motion affinities throughout the whole video using epipolar geometry and the motion information provided by optical flow. Finally, co-regularized multi-view spectral clustering is used to fuse the two affinity matrices and obtain the final clustering. Our method shows state-of-the-art performance on the KT3DMoSeg dataset, which contains complex motions and scene structures. Being able to identify moving objects allows us to remove them for map building when using visual SLAM or SFM.

\end{abstract}

\section{INTRODUCTION}

The objective of motion segmentation is to divide a video frame into regions segmented by common motions. Motion segmentation from a moving camera is particularly important as it has various applications in areas like autonomous navigation, robotics and SLAM. In a dynamic scene, the video camera is moving at an unknown velocity with respect to the environment. Such scenarios pose many challenges to motion segmentation methods such as motion degeneracy, motion parallax and motion on the epipolar plane \cite{Hartley2004}. Existing monocular motion segmentation methods often fail when facing these challenges \cite{xi_multi-motion_2022, xu_motion_2015, xu_motion_2018}, or need specific assumptions on the scene structure, object classes or motion types \cite{mohamed_monocular_2021, vertens_smsnet_2017, ramzy_rst-modnet_2019}. 

In order to overcome these limitations and achieve high quality instance motion segmentation results regardless of scene structures and motion types, we draw our inspiration from two branches of well studied motion segmentation approaches: optical flow based methods and point trajectory based methods. These two types motion cues are not only complementary in nature (long-term vs instantaneous motion), but they can also be used to derive highly complementary geometric and motion models for different motion types and scene structures: points trajectory based methods, when analyzed using epipolar geometry, will fail if the motion is mainly on the epipolar plane, but it is robust to depth variations, perspective effects and motion parallax; on the other hand, optical flow based methods will fail on these challenges, but it is robust to motions on the epipolar plane. We propose to combine these two complimentary motion cues at object level to obtain a robust and comprehensive motion representation of the scene. By using the state-of-the-art methods for object recognition, detection, segmentation and tracking, we can easily obtain an objectiveness prior (i.e., an initial grouping) of all motion data. This approach enables us to analyze motions of each individual object, which is crucial for both robots and humans to build their situational awareness and scene understanding capabilities \cite{jiang_what_2021, bavle_slam_2023}. 

To build our motion segmentation framework, we first leverage state-of-the-art deep learning models \cite{rajic_segment_2023, liu_grounding_2023, zhang_recognize_2023, yang_decoupling_2022} to recognize, detect, segment and track any common objects throughout the video. Then, for every object in the video, we obtain (1) a set of sparse point trajectories for each object and (2) a dense optical flow mask for each object on each frame. By using object-specific sparse point trajectories and optical flow masks, we are able to derive object-specific geometric models (i.e. fundamental matrices based on epipolar geometry) and instantaneous motion models respectively on every frame pair where the object is visible. By fusing these two highly complementary models, we are theoretically able to model the vast majority of motions even in complex scenes. Our experiments show significant improvements on motion segmentation results in challenging scenarios, highlighting the approach's potential in real-world applications. In summary, the key contributions of our paper are as follows: 
\begin{enumerate}
    \item We combine the well-studied fundamental matrix motion model and the optical flow based instantaneous motion model using multi-view spectral clustering to model multiple complex motions in challenging scenes.
    \item We show how to model different motions at object level by incorporating per-frame objectiveness prior obtained from recent computer vision foundational models.
    \item We achieve state-of-the-art result on the challenging KT3DMoSeg dataset \cite{xu_motion_2018} in terms of both producing high-quality point trajectory clustering and pixel-level masks for individual moving instances. 
\end{enumerate}

\section{RELATED WORK}

Monocular motion segmentation can be broadly categorized into three groups: (1) Intensity based methods \cite{negahdaripour_direct_1987, sekkati_variational_2007, wedel_structure-_2009, leibe_its_2016, bideau_best_2018}, (2) sparse correspondence based methods \cite{delong_fast_2010, isack_energy-based_2012, hutchison_object_2010, brox_object_2010, elhamifar_sparse_2013, ochs_segmentation_2014, lai_motion_2017, xu_motion_2018, vedaldi_usage_2020} and (3) deep learning methods \cite{vertens_smsnet_2017, siam_modnet_2018, bosch_deep_2021, ramzy_rst-modnet_2019, dave_towards_2019, cao_learning_2019, faisal_epo-net_2020, mohamed_monocular_2021, meunier_em-driven_2023}. 

\subsection{Intensity Based Methods}
Intensity based methods can be further categorized into indirect and direct methods. Indirect methods \cite{sekkati_variational_2007, wedel_structure-_2009, leibe_its_2016} rely on pixel-wise correspondences as input, and produce a pixel-wise segmentation mask indicating different motion groups. Such input is usually optical flow, which assumes the brightness or color intensity of every point in the scene remains the same throughout the whole sequence. In contrast, direct methods \cite{negahdaripour_direct_1987} do not require optical flow -- they combine the two processes of optimizing for the brightness consistency constraint and estimating the motion models together and directly take a pair of images as input. Most recent works on intensity based methods use optical flow based indirect methods, possibly due to 
the fast advance in optical flow estimation \cite{teed_raft_2020, sun_disentangling_2022}. Such methods usually adopt an iterative optimization approach to estimate the motion model and motion region simultaneously.

Intensity-based methods tend to perform well on scenes without strong depth variations or motion parallax as the motion flow vectors projected to a 2D image from the 3D space are determined by both the depth and the screw motion of objects \cite{mitiche_computer_2014}. Therefore, if the scene contains strong depth variation (e.g. road scenes), intensity-based methods will fail to distinguish if a part of the image is moving independently or is just at a different depth from its surroundings.

\subsection{Sparse Correspondence Based Methods}
Sparse correspondence based methods can be further categorized into two-frame and multi-frame methods. Two frame methods \cite{delong_fast_2010, isack_energy-based_2012} usually recover motion parameters by solving an iterative energy minimization problem of finding a certain number of geometric models (e.g., fundamental matrices) on a set of matched feature points, to minimize an energy function that evaluates the quality of the overall clustering of correspondences. Multi-frame methods often operate on manually corrected trajectory points obtained from a dense optical flow tracker. Such methods usually perform clustering on affinity matrices constructed using the results of geometric model fitting \cite{lai_motion_2017, xu_motion_2018, vedaldi_usage_2020} or pairwise affinities derived from spatio-temporal motion cues and appearance cues \cite{brox_object_2010, ochs_segmentation_2014}. 

Point trajectories, when analyzed using epipolar geometry, are robust to depth variations in the scene, but are prone to motions on the epipolar plane. \cite{vedaldi_usage_2020} uses trifocal tensor as a more robust model to analyze point trajectories. Trifocal tensor is more robust to noise and motions on the epipolar plane, but it is harder to optimize and prone to failure when the three cameras are close to being colinear \cite{Hartley2004}, which can often happen on road scenes. \cite{xu_motion_2018, xi_multi-motion_2022} produce promising results by combining multiple geometric models, but still fails to produce a coherent and consistent segmentation on some scenes. Methods using spatio-temporal information and appearance cues \cite{brox_object_2010, ochs_segmentation_2014} suffer from similar issues. In addition, although they tend to perform a bit better than geometric methods on motions with less rigidity \cite{lezama_track_2011}, they perform worse on scenes with motion parallax or strong camera motions. 

\subsection{Deep Learning Based Methods} 
Deep learning based methods usually takes a pair or a sequence of input frames as input and directly produces a either a binary segmentation mask of moving vs static objects \cite{vertens_smsnet_2017, siam_modnet_2018, bosch_deep_2021}, or a multi-label segmentation mask showing different objects of different motions \cite{ramzy_rst-modnet_2019, dave_towards_2019, cao_learning_2019, faisal_epo-net_2020, mohamed_monocular_2021, meunier_em-driven_2023}. Most deep learning methods use CNNs amd their network architecture can be broadly summarized to have the following main components: (1) a module to extract the appearance information from consecutive frames, (2) a module to extract motion information from the same frames,  (3) a module to fuse the appearance and motion information, and (4) a decoder to generate the final segmentation. These methods are usually fully-supervised and require a large amount of training data. Besides, these methods are often only able to perform binary motion detection (moving vs. static) \cite{ramzy_rst-modnet_2019, faisal_epo-net_2020, bosch_deep_2021}, or they are only able to detect instance motions for specific scenes and limited number of object types they are trained on \cite{cao_learning_2019, mohamed_monocular_2021}

\section{METHODOLOGY}

\begin{figure*} [thbp]  
    \centering
    \includegraphics[width=\textwidth]{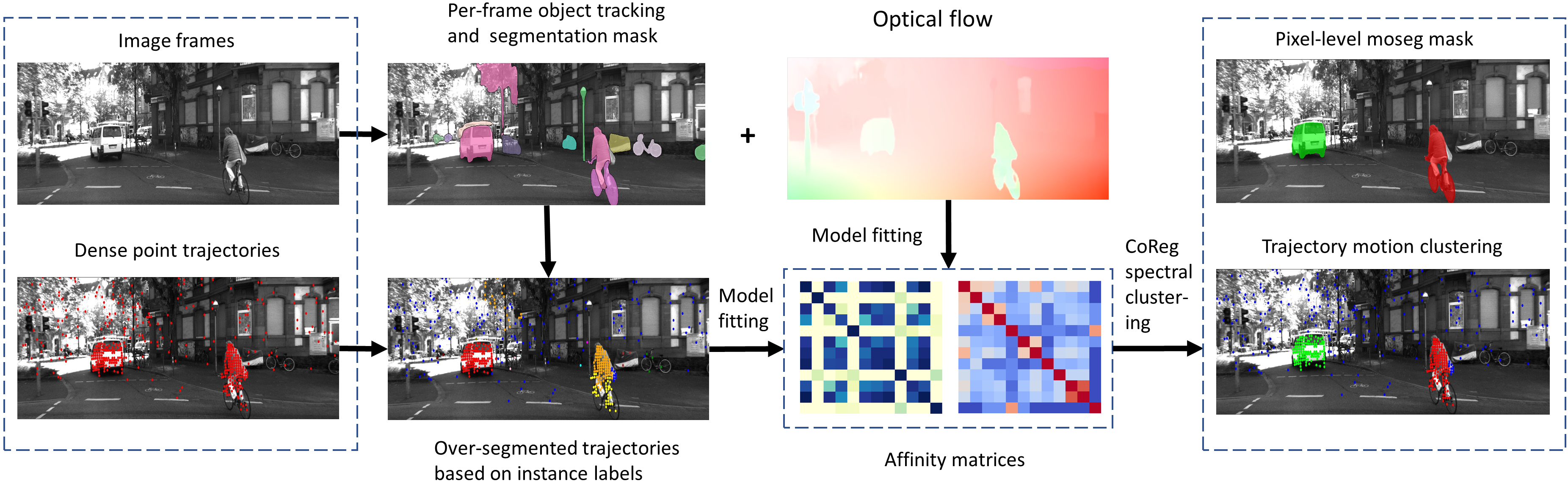} 
    \caption{Motion Segmentation Pipeline}
    \label{fig: Motion Segmentation Pipeline}
\end{figure*}

\subsection{Object Recognition, Detection, Segmentation and Tracking}
Figure \ref{fig: Motion Segmentation Pipeline} shows a diagram of our motion segmentation pipeline. In order to identify all motions in a video sequence at object level, we first identify every common object in the video and track their movements throughout the video. We achieve this by using the most recent foundational models in object recognition (Recognize Anything Model)\cite{zhang_recognize_2023}, detection (Grounding DINO model) \cite{liu_grounding_2023} and segmentation (Segment Anything Model) \cite{rajic_segment_2023}, and a state-of-the-art object tracker (DeAOT) \cite{yang_decoupling_2022}. We adapt our preprocessing pipeline from Segment and Track Anything (SAMTrack) \cite{cheng_segment_2023}, which is an object segmentation and tracking framework made of the Grounding DINO model, Segment Anything Model (SAM) and the DeAOT tracker. SAMTrack allows the user to segment and track any specific objects in the video with a text prompt. To make our system fully automatic and universal to all motions in most scenes, we avoid using the user-defined text prompt by adding RAM at the beginning of our pipeline to automatically recognize any common objects in the video. In summary, our whole preprocessing pipeline consists of the following main steps: 1) Use RAM to recognize any common objects in the first frame of the video; 2) Feed the output of RAM as a text prompt to the Grounding DINO model to obtain object bounding boxes; 3) Feed these bounding boxes to SAM to obtain an instance segmentation mask of the first frame. Non-max suppression was used to remove objects with an IoU score $>$ 0.5 and instances whose bounding boxes are larger than half the image are also removed; 4) Use the DeAOT tracker to track each object's mask throughout the entire video. In order to detect new objects entering the scene in the middle of the video, we run step 1) every {\it l} frames to check if there are new objects. If so, we then run steps 2) to 4) to segment and track the new objects together with the existing objects in the frame. The number {\it l} is video-specific, for example, more dynamic videos with more objects entering and leaving the scene would benefit from a smaller {\it l}. 

\subsection{Obtaining Point Trajectories and Optical Flow Masks}
Once we have the instance segmentation mask for every frame of the video, we then need to obtain a set sparse of point trajectories and a dense optical flow mask on every object at every frame where it's visible. Essentially, we aim to obtain instance point trajectories and instance optical flows. We use one of the state-of-the-art methods \cite{sun_disentangling_2022-1} to obtain optical flow for all frames, then label each flow vector with the object label of the pixel. 
Ideally, point trajectories need to be sampled from each object and automatically tracked at each frame using a point tracker (e.g., \cite{harley_particle_2022, doersch_tapir_2023}), however, for benchmarking purposes, we use the manually corrected point trajectories provided by the KT3DMoSeg dataset. We assign every point trajectory with an initial object label using the per-frame instance segmentation masks. If a trajectory does not belong to any object, we label it as the background. Due to inaccuracy in instance segmentation and tracking, a point trajectory can be identified to be on different objects or background in different frames. In such case, we assign its object label to be the most frequent label it is identified as. In the future, we will use a point tracker as well as robust point sampling and occlusion handling techniques to automatically generate point trajectories for all detected objects.

\subsection{Model Fitting}
After obtaining instance point trajectories and optical flow masks for each frame, we compute the motion models of each object to model its motion throughout the video. To compute the epipolar geometry based motion models using point trajectories, we compute a fundamental matrix of each object between every {\it f} frames by solving ${p'T F p = 0}$ using the eight-point algorithm. If a degenerate case is encountered for the fundamental matrix, we do not use it. For the optical flow based motion model, we use a full quadratic motion model with 12 parameters to model the instantaneous object screw motion:
\begin{equation} \label{eq:1}
\begin{split}
f(x, y) = & (a + b x + c y + d x^2 + e xy + f y^2, \\
& g + h x + i y + j x^2 + k xy + l y^2)
\end{split}
\end{equation}

where {\it (x, y)} is the 2D coordinates of the pixels relative to the image center. Since we already have the instance optical flow field, we can obtain the following equation:
\begin{equation} \label{eq:2}
\begin{split}
f(x, y) = (u, v)
\end{split}
\end{equation}

where $(u, v)$ is the optical flow vector of the pixel. We fit the function \ref{eq:1} above to the the optical flow vectors of every object and solve for the 12 parameters representing the object motion model by optimizing the mean squared error. We use this specific motion model as it's a simplified version of the classic Longuet-Higgins and Pruzdny model equations \cite{longuet-higgins_interpretation_1980}, which model's the instantaneous screw motion of rigid objects at arbitrary depth. Since it's not possible to solve for the depth of each pixel, this motion model assumes the objects' depths are only slightly different. It was shown to perform well on scenes with limited motion parallax \cite{meunier_em-driven_2023}, nevertheless, it often fails when there is strong motion parallax and depth variations.

\subsection{Constructing Affinity Matrices}
After all fundamental matrices and optical flow motion models are computed, each object will have a fundamental matrix between every {\it f} frames and an optical flow motion model between every two frames. By fitting every object's trajectory points and optical flow vectors to every other object's fundamental matrix and optical flow motion model on the same frame pair, we can obtain the residuals of every object to all other objects' motion models respectively. We use Sampson distance as the residual for fundamental matrix \cite{Hartley2004} and mean squared error for optical flow motion model. Assuming there are {\it k} objects in the scene, for the {\it i}-th object  at the {\it m}-th frame pair, we obtain the following residual vectors under the fundamental matrix and optical flow motion models:
\begin{equation*} \label{eq:3}
\setlength{\jot}{10pt} 
\begin{split}
{\pmb r_{o}}_i^m = [{r_{o}}_{i,1}^m, {r_{o}}_{i,2}^m, ..., {r_{o}}_{i,k}^m], 
\\
{\pmb r_{f}}_i^m = [{r_{f}}_{i,1}^m, {r_{f}}_{i,2}^m, ..., {r_{f}}_{i,k}^m]
\end{split}
\end{equation*}

where ${r_{o}}_{i,k}^m$ is the mean residual for fitting the parametric motion model of object {\it i} on the optical flow vectors of object {\it k} between frames {\it m} and {\it m} + 1, and ${r_{f}}_{i,k}^m$ is the mean Sampson error for fitting the fundamental matrix of object {\it i} on the trajectory points of object {\it k} between frames {\it m} and {\it m + f}. We construct two affinity matrices encapsulating the pairwise motion affinities between each pair of objects using the ordered residual kernal (ORK) \cite{chin_ordered_2009}. More specifically, for each object, we sort its residual vectors in ascending order and define a threshold to select the smallest {\it t}-th residual as inliers. We define ${\pmb c_{i}} = \{0, 1\}^K$ as the inlier mask to denote if an object i is an inlier for each of the {\it K} motion models, and the pairwise motion affinity between objects {\it i} and {\it j} can be computed as ${\pmb a_{ij} = \pmb c_{i}^\intercal \pmb c_{j}}$, which denotes the co-occurrence between two objects as an inlier of all motion models. ORK is robust to outliers and makes the affinity matrix more adaptive to different scenes by reducing the need to set scene specific inlier thresholds.

\subsection{Co-Regularized Multi-view Spectral Clustering}
After constructing the affinity matrices, we use the epsilon-neighborhood scheme \cite{lai_motion_2017} to sparsify the affinity matrices. We adapt co-regularized multi-view spectral clustering \cite{kumar_co-regularized_2011} to fuse the two affinity matrices together to obtain the final clustering of object motions and trajectory points. Co-regularized multi-view spectral clustering uses an regularization term to encourage consensus between different views and is shown to perform well on fusing multiple geometric models for a consistent representation of data \cite{xu_motion_2018}.

\section{PRELIMINARY RESULTS \& CONCLUSIONS}

\begin{figure} [htb] 
    \centering
    \begin{subfigure}{0.48\textwidth}
        \includegraphics[width=\linewidth]{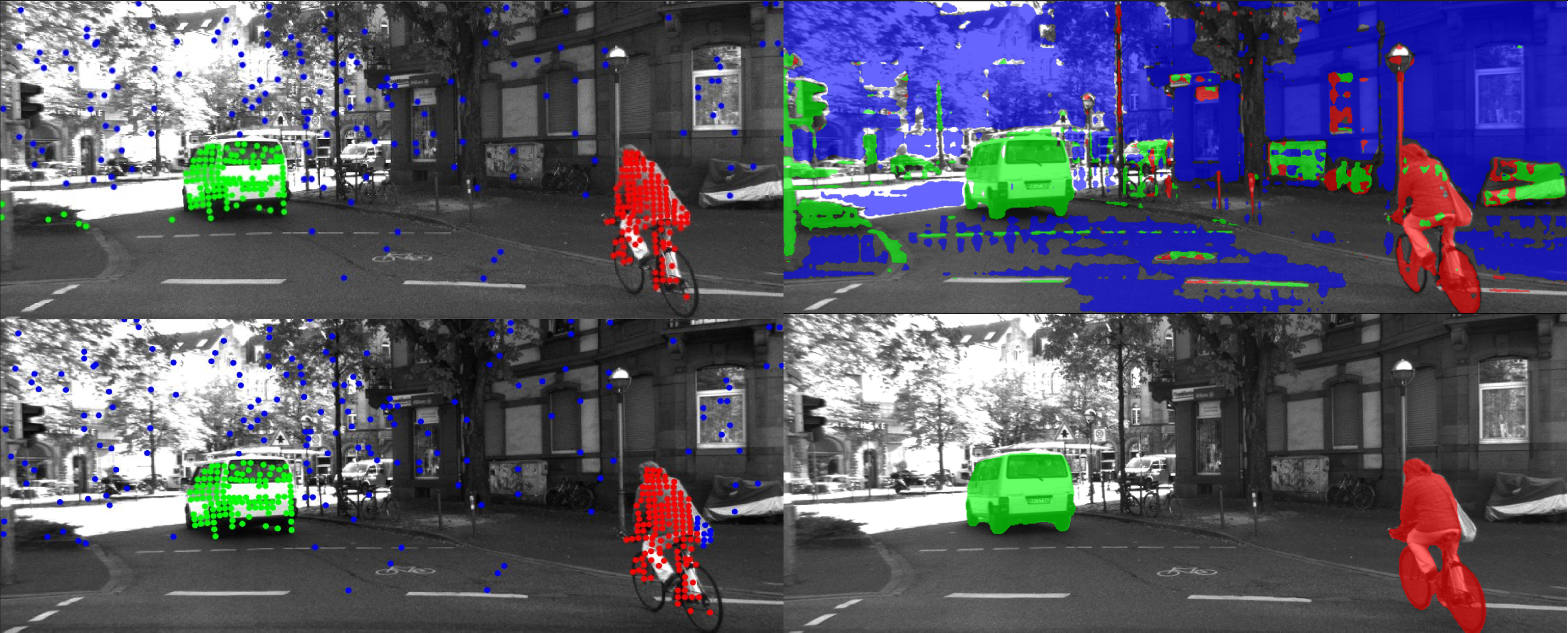}
    \end{subfigure}
    \caption{Trajectory clustering results and the generated segmentation masks by \cite{xu_motion_2018} (top) and our method (bottom)}
    \label{fig: Qualitative Segmentation Results}
\end{figure}

\begin{figure} [htb] 
    \centering
    \begin{subfigure}{0.48\textwidth}
        \includegraphics[width=\linewidth]{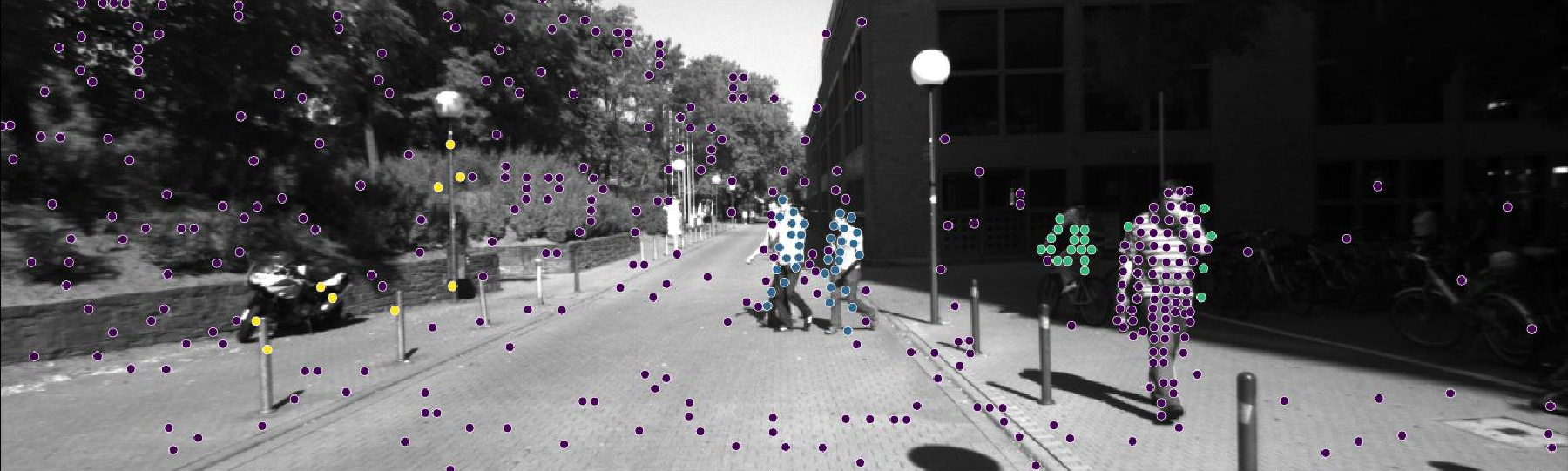}
    \end{subfigure}
    \vfill
    \begin{subfigure}{0.48\textwidth}
        \includegraphics[width=\linewidth]{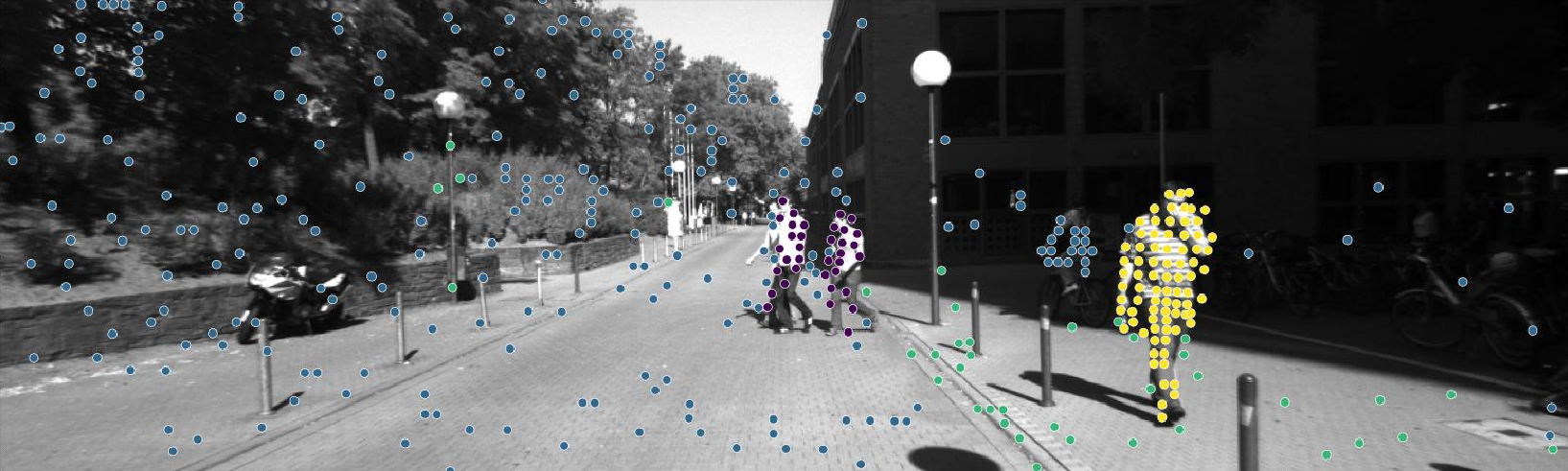}
    \end{subfigure}
    \vfill
    \begin{subfigure}{0.48\textwidth}
        \includegraphics[width=\linewidth]{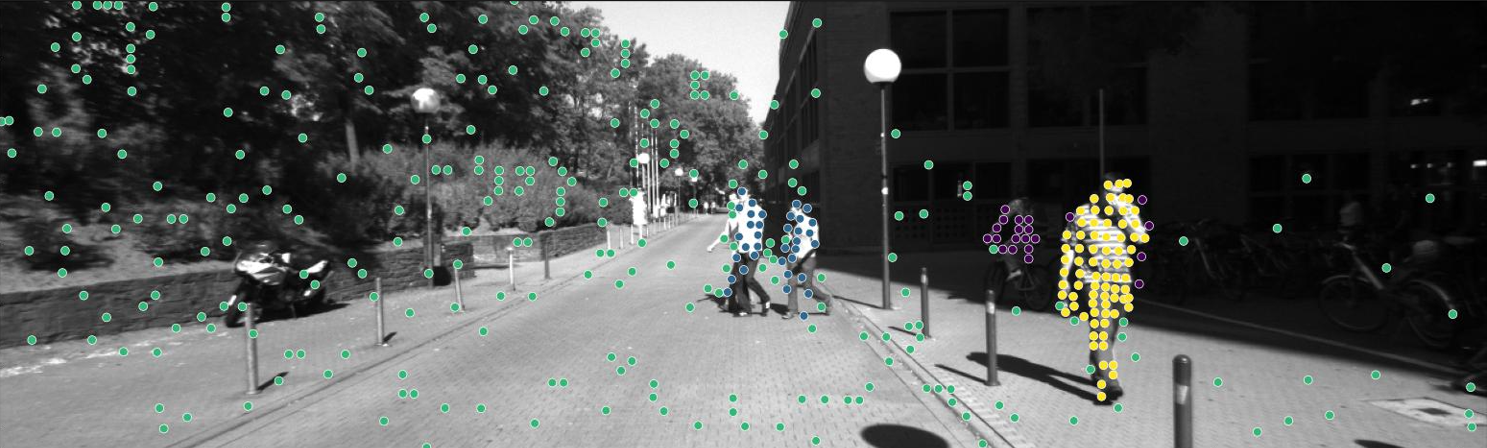}
    \end{subfigure}
    \vfill
    \begin{subfigure}{0.48\textwidth}
        \includegraphics[width=\linewidth]{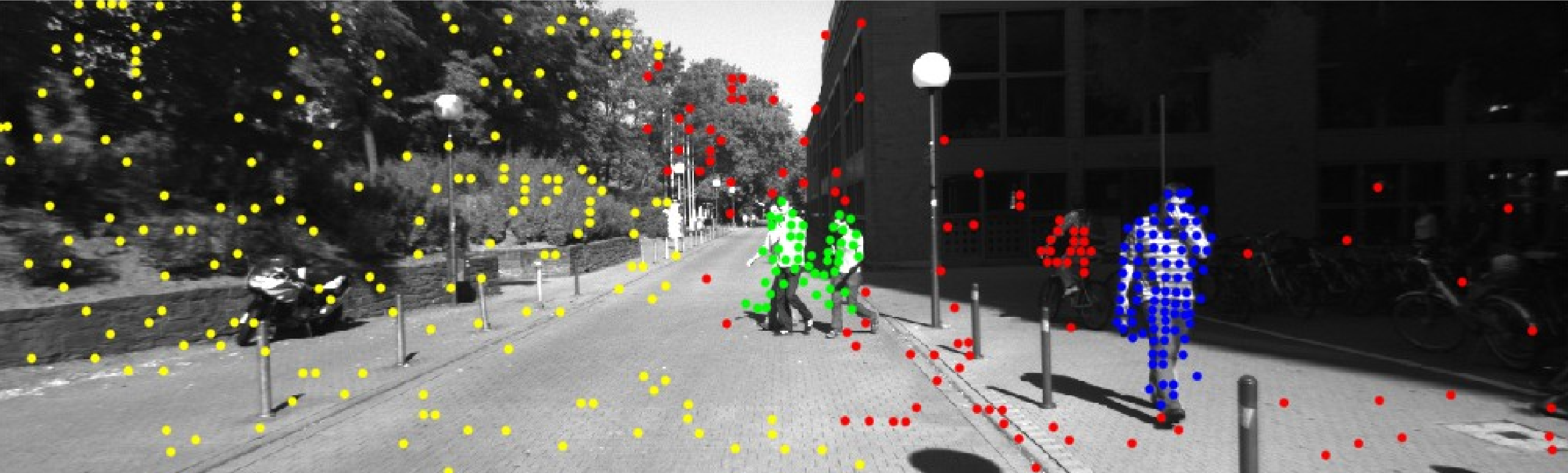}
    \end{subfigure}
    \vfill
    \begin{subfigure}{0.48\textwidth}
        \includegraphics[width=\linewidth]{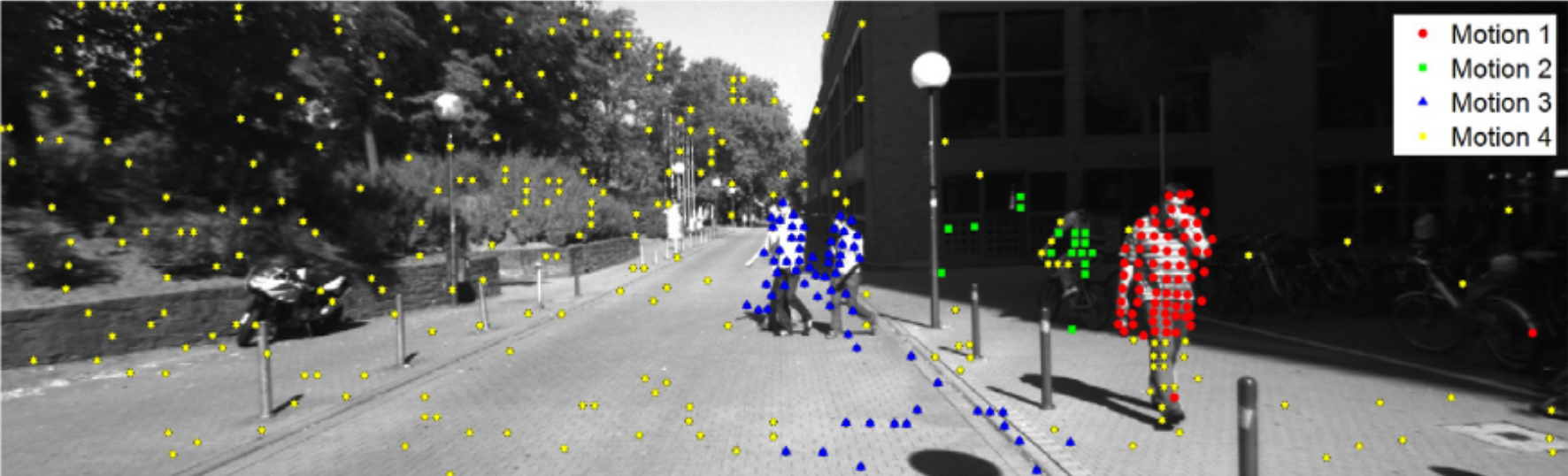}
    \end{subfigure}
    \caption{Qualitative ablation study and comparison with state-of-the-art methods on the Seq38\_Clip02 sequence. From top to bottom: fundamental matrix only, optical flow only, fundamental matrix + optical flow, MVC \cite{xu_motion_2018}, CMFO \cite{xi_multi-motion_2022}}
    \label{fig: Qualitative Trajectory Clustering Results}
\end{figure}

We tested our method on the KT3DMoSeg dataset since it is the only existing dataset involving challenging scenes and multiple complex motions. Since KT3DMoSeg uses pre-defined point trajectories, it could occur that an object segmented by our preprocessing pipeline has fewer than 7 point trajectories. In this case, it's not possible to compute fundamental matrices for the object, but we can still obtain its residual vectors by fitting its trajectory points (if there is any) and the optical flow vectors on the motion models of other objects to compute its pairwise motion affinity scores. Figure \ref{fig: Qualitative Segmentation Results} shows a qualitative comparison between the segmentation masks generated by our method and a baseline we created using a state-of-the-art method \cite{xu_motion_2018} whose code is publicly available. To establish the baseline, we prompt SAM using the clustered trajectories of \cite{xu_motion_2018} (top left) to produce the segmentation mask. Even though \cite{xu_motion_2018} achieves very low clustering error rate on this sequence, the quality of the generated segmentation mask is still worse than ours since a few wrongly labeled trajectory points can easily mislead SAM. It's also not able to recognize which motion is background. Figure \ref{fig: Qualitative Trajectory Clustering Results} shows qualitative ablation study and comparison to state-of-the-art methods. Table \ref{tab: Quantitative Results} shows quantitative results comparing to existing methods. Our method outperforms the state-of-the-art both qualitatively and quantitatively. We also compare the total running time (measured on an Intel 13900K CPU and an NVIDIA RTX 4090 GPU) of our method with \cite{xu_motion_2018} for reference. Our method is more than twice as slow since it's not yet optimized.

\begin{table}[h]
    \centering
    \caption{Quantitative results in terms of average classification error (\%) and total running time (s). Lower is better}
    \label{tab: example}
    \begin{tabular}{ccc}
        \toprule
        \textbf{Methods} & \textbf{Avg. Error Rate (\%)} & \textbf{Running Time (s)}\\
        \midrule
        LSA \cite{yan_general_2006} & 38.30 & -\\ 
        GPCA \cite{vidal_multiframe_2008} & 34.60 & -\\
        ALC \cite{rao_motion_2010} & 24.31 & -\\
        SSC \cite{elhamifar_sparse_2013} & 33.88 & -\\
        LRR \cite{liu_robust_2013} & 33.67 & -\\
        MVC \cite{xu_motion_2018} & 10.99 & 1409.6\\
        CMFO \cite{xi_multi-motion_2022} & 6.73 & -\\
        Ours & \textbf{5.78} & 3230.1\\
        \bottomrule
    \end{tabular}
    \label{tab: Quantitative Results}
\end{table}

\clearpage
\bibliography{IEEEexample}
\bibliographystyle{IEEEtran}

\end{document}